\documentclass[a4paper,fleqn]{cas-dc}
\usepackage{float}
\usepackage{color}
\usepackage[numbers,sort&compress]{natbib}
\usepackage{ragged2e}
\usepackage[font={sf,bf},textfont=md]{caption}
 
\def\tsc#1{\csdef{#1}{\textsc{\lowercase{#1}}\xspace}}
\tsc{WGM}
\tsc{QE}
\tsc{EP}
\tsc{PMS}
\tsc{BEC}
\tsc{DE}

\usepackage{marginnote}

\begin{document}
	\begin{sloppypar}
		\let\WriteBookmarks\relax
		\def\floatpagepagefraction{1}
		\def\textpagefraction{.001}
		\shorttitle{Hongbo Bi et~al./Elsevier xxx (xxxx) xxx}
		\shortauthors{Hongbo Bi et~al.}
		
		\title [mode = title]{Towards Accurate RGB-D Saliency Detection with Complementary Attention and Adaptive Integration}
		%
		%

		\author[1]{Hong-Bo Bi}[
		style=chinese,
		]
		
		
		\address[1]{School of Electrical Information Engineering, Northeast Petroleum University, Daqing 163000, China.}
		
		\author[1]{Zi-Qi Liu}[style=chinese]
		
		\author[1]{Kang Wang}[
		style=chinese]
		
		
		\address[2]{Department of Optical Electrical and Computer Engineering, University of Shanghai for Science and Technology, Shanghai 200093, China}
		
		\author[2]{Bo Dong}[style=chinese]
		\author[3]{Geng Chen}[style=chinese]
		\address[3]{Inception Institute of Artificial Intelligence, Abu Dhabi, UAE}
		
		\author[4]{Ji-Quan Ma}[style=chinese]
		\address[4]{Department of Computer Science and Technology, Heilongjiang University, Harbin 150080, China}
		\begin{abstract}
			Saliency detection based on the complementary information from RGB images and depth maps has recently gained great popularity.
			In this paper, we propose \textbf{C}omplementary \textbf{A}ttention and \textbf{A}daptive \textbf{I}ntegration \textbf{Net}work (\textbf{CAAI-Net}), a novel RGB-D saliency detection model that integrates complementary attention based feature concentration and adaptive cross-modal feature fusion into a unified framework for accurate saliency detection.
			Specifically, we propose a context-aware complementary attention (CCA) module, which consists of a feature interaction component, a complementary attention component, and a global-context component.
			The CCA module first utilizes the feature interaction component to extract rich local context features. 
			The resulting features are then fed into the complementary attention component, which employs the complementary attention generated from adjacent levels to guide the attention at the current layer so that the mutual background disturbances are suppressed and the network focuses more on the areas with salient objects.
			Finally, we utilize a specially-designed adaptive feature integration (AFI) module, which sufficiently considers the low-quality issue of depth maps, to aggregate the RGB and depth features in an adaptive manner.
			Extensive experiments on six challenging benchmark datasets demonstrate that CAAI-Net is an effective saliency detection model and outperforms nine state-of-the-art models in terms of four widely-used metrics.
			In addition, extensive ablation studies confirm the effectiveness of the proposed CCA and AFI modules.
		\end{abstract}
		
		
		

		\maketitle
		
		\section{Introduction}
		Salient object detection (SOD), which segments the most attractive objects in an image, has drawn increasing research efforts in recent years~\cite{DBLP:journals/cvm/BorjiCHJL19,DBLP:journals/tip/BorjiCJL15,DBLP:journals/pami/ChengMHTH15,zhao2019egnet,fu2019deepside:,DBLP:conf/iccv/SuLZXT19,liu2020deep,DBLP:conf/cvpr/WangSC019,DBLP:journals/pami/WangSDBY20,dong2021bcnet}.
		SOD has a large number of applications, such as object recognition~\cite{Alex2020Adversarial}, image video compression~\cite{Norwahidah2020Analysis}, image retrieval~\cite{gao20123-d,Kamarasan2020Quadhistogram}, image redirection~\cite{Bayrock2002Image}, image segmentation~\cite{Grady2006Random,BRINet_CVPR2020}, image enhancement~\cite{Osher1990Feature}, quality assessment~\cite{Bansiya2002A}, etc.
		With the rapid progress in this field, a number of derived techniques are developed. Typical instances include video saliency detection~\cite{DBLP:journals/tip/WangSS18,DBLP:journals/tip/ChenLWQH17,DBLP:journals/spl/ChenLLQH18,DBLP:journals/tip/WangSS15,DBLP:journals/tip/ChenWPZQ20,DBLP:conf/cvpr/WangSGCB18,DBLP:journals/pami/WangSYP18}, co-saliency detection~\cite{Hongbo2020,deng2020re}, stereo saliency detection~\cite{DBLP:journals/tvcg/WangSYM17}, etc.
		
		The perception of depth information is the premise of human stereoscopic vision. Therefore, considering depth information in SOD can better imitate the human visual mechanism and improve the detection accuracy.
		In recent years, increasing research effort has been made to study the RGB-D saliency detection \cite{DBLP:conf/cvpr/LiY15,DBLP:journals/jvcir/DingLHSW19,chen2020improved,Zhengyi2020A,li2020icnet:,zhai2020bifurcated,Fu2020JLDCF,Zhang2020Bilateral,chen2020ef,huang2020multi}.
		Existing methods employ different schemes to handle the multi-level multi-modal features. For the multi-level features, Liu et~al.~\cite{Nian2020PiCANet} utilized pixel-wise contextual attention network to focus on context information for each pixel and hierarchically integrate the global and local context features. Wang et al.~\cite{DBLP:conf/cvpr/WangZSHB19} devised a pyramid attention structure to concentrate more on salient regions based on typical bottom-up/top-down network architecture. Zhang et al.~\cite{Zhang2017Amulet} developed an aggregating multi-level convolutional feature framework to extract the multi-level features and integrate them into multiple resolutions. For the fusion of the multi-modal features, Liu et al.~\cite{liu2020multi-level} took depth maps as the fourth channel of the input and employed a parallel structure to extract features through spatial/channel attention mechanisms.
		Piao et al.~\cite{piao2020A2dele} exploited a multi-level cross-modal way to fuse the RGB and depth features, and proposed a depth distiller to transfer the depth information to the RGB stream.
		Li et al.~\cite{li2020icnet:} designed an information conversion module to fuse high-level RGB and depth features adaptively, and RGB features at each level were enhanced by weighting depth information.
		Piao et al.~\cite{Yongri2020Depth} adopted a depth refinement block based fusion method for each level RGB and depth features. More details can be found in the recently released RGB-D survey and benchmark papers~\cite{fan2019rethinking,zhou2020rgbd,DBLP:journals/corr/abs-1904-09146}.
		
\begin{figure}[tb]
			\begin{center}
				\includegraphics[width=8.2cm]{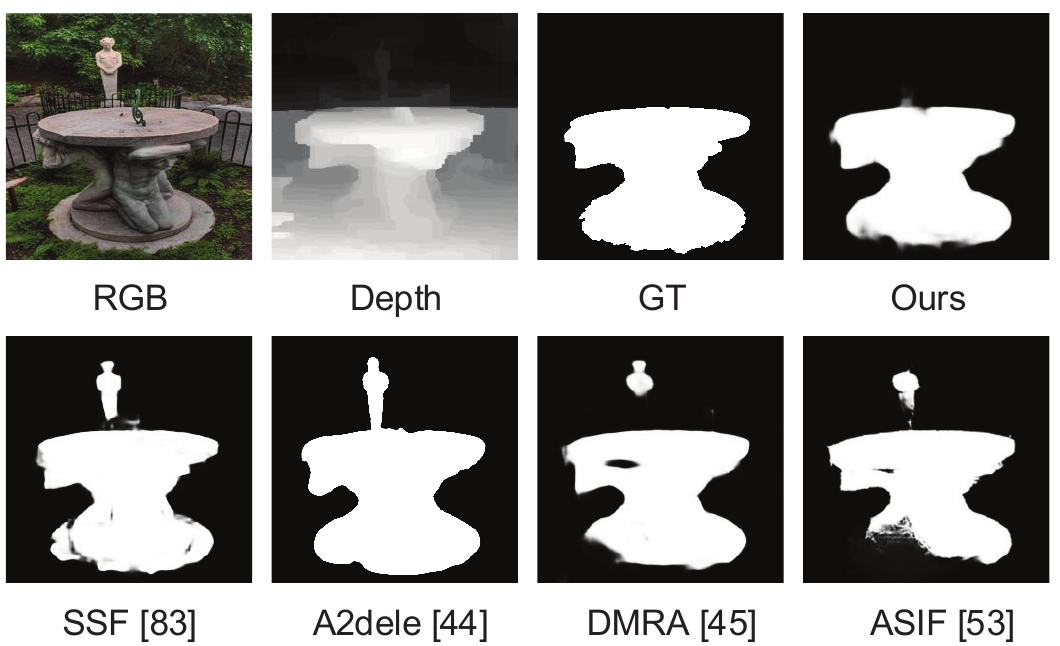}
			\end{center}
			\vspace{-12pt}
			\caption{Saliency maps of state-of-the-art deep-based RGB-D models in a complex scene that is locally similar.}
			\label{fig:compare}
		\end{figure}

		Despite their advantages, most existing deep-based RGB-D saliency detection methods suffer from two major limitations. 
		{First, although attention mechanisms have been adopted, most existing methods only rely on a kind of attention mechanisms, e.g., channel attention, spatial attention, etc. This results in the drawback that the network is unable to sufficiently explore and make full use of the attention for improving the performance.}
		Second, existing methods usually overlook the noise nature of depth maps, and directly fuse the RGB and depth features by simple concatenation or addition.
		More reasonable fusion of multi-level and cross-modal features can effectively reduce the error rate caused by misidentification. This is particularly important for the salient object detection in the interference environment, e.g., complex, low-contrast, similar background, etc.
		As shown in Fig.~\ref{fig:compare}, the low-quality depth information and locally similar scene affects the performance of existing cutting-edge models, making them unable to accurately detect the salient objects.
		
		To address these limitations, in this paper, we propose a novel RGB-D saliency detection model, called \textbf{C}omplementary \textbf{A}ttention and \textbf{A}daptive \textbf{I}ntegration \textbf{Net}work (\textbf{CAAI-Net}), which employs a complementary attention mechanism along with adaptive feature fusion to detect mesh saliency from multi-modal RGB-D images.
		Our CAAI-Net effectively resolves the drawbacks in existing methods with a more comprehensive attention mechanism and a novel fusion strategy, which considers the low-quality issue of depth maps and fuses multi-modal features in an adaptive manner.
		Specifically, we employ two backbones to extract multi-level features from RGB images and depth maps.
		The multi-level features are first divided into low-level and high-level features according to their locations in the backbones.
		For the low-level features, the semantic information of the different channels is almost indistinguishable, therefore we adopt spatial attention (SA) components to refine the features rather than using channel attention (CA) components.
		The attention component is employed to suppress the useless background information and locate the informative features.
		For the high-level features, we propose a context-aware complementary attention (CCA) module for better informative feature concentration and noisy feature reduction.
		The CCA module consists of a feature interaction component, a complementary attention component, and a global-context component.
		The feature interaction component is designed to extract the local context features using a pyramid structure, which supplements missing information from adjacent levels.
		The resulting features are then fed to the complementary attention component, which is a mixture of CA and SA components with effective inter-level guidance.
		In addition, the global-context component further supplements the details.
		Finally, we design an adaptive feature integration (AFI) module to adaptively fuse the cross-modal features at each level.
		The AFI module employs the fusion weights generated from the adjacent levels as guidance to obtain enhanced RGB features, and then fuse the enhanced RGB and depth features in an adaptive manner.
		
		In summary, our contributions lie in three-fold:
		\begin{itemize}
			\item{{We propose the CCA module, which is able to extract the informative features highly related to the accurate saliency detection.
					In the CCA module, the feature interaction component employs a pyramid structure along with nested connections to extract rich context features.
					The complementary attention component refines the features to capture highly informative features, while effectively reducing the noisy feature disturbances.
					The global-context component supplements the details to enrich the features.}}
			\item{We propose a novel adaptive feature fusion module, AFI, which adaptively integrates the multi-modal features at each level.
				The AFI module is able to self-correct the ratio of different feature branches.
				Moreover, the feature coefficients automatically generated from pooling and softmax layers are assigned to the enhanced RGB features and depth features to balance their contributions to the feature fusion.} 
			\item{Extensive experiments on six benchmark datasets demonstrate that our CAAI-Net outperforms nine state-of-the-art (SOTA) RGB-D saliency detection methods, both qualitatively and quantitatively. In addition, the effectiveness of the proposed modules is validated by extensive ablation studies.}
		\end{itemize}
		
		Our paper is organized as follows. In Section~\ref{sec:related_work}, we will introduce related work. In Section~\ref{sec:method}, we will describe our CAAI-Net in detail. In Section~\ref{sec:experiments}, we will present the datasets, experimental settings, and results. Finally, we will conclude our work in Section~\ref{sec:conclusion}.

		\section{Related Works}\label{sec:related_work}
		In this section, we discuss a number of works that are closely related to ours. These works are divided into three categories, including RGB-D saliency detection, global context and local context mechanism, and attention mechanism.

		\subsection{RGB-D Saliency Detection}
		The early RGB-D saliency detection methods are mostly based on hand-crafted features, such as color~\cite{Borji2015What}, brightness~\cite{DBLP:conf/cvpr/JiangWYWZL13}, and texture~\cite{Yang2013Saliency}.
		However, these methods are unable to capture the high-level semantic information of salient objects and have low confidence level and low recall rate.
		Afterwards, deep convolutional neural network (CNN) is introduced and has shown remarkable success in RGB-D saliency detection.
		Zhou et al.~\cite{DBLP:journals/ivc/ZhouLGLZ20} utilized multi-level deep RGB features to combine the attention-guided bottom-up and top-down modules, which is able to make full use of multi-modal features.
		Li et al.~\cite{Chongyi2020ASIF} proposed an attention steered interweave fusion network to fuse cross-modal information between RGB images and corresponding depth maps at each level.
		These methods utilize attention modules to improve the ability of acquiring local information for salient objects detection.
		Some of them consider spatial attention mechanism, while others use channel attention mechanism to guide RGB-D saliency detection.
		In our work, we take full advantage of both attention mechanisms for improved performance.
		
		A number of RGB-D saliency detection methods focus on the fusion of cross-modal information.
		Xiao et al.~\cite{DBLP:journals/access/XiaoLPCHG20} employed a CNN-based cross-modal transfer learning framework to guide the depth domain feature extraction.
		Wang et al.~\cite{DBLP:journals/access/WangG19} designed two-streamed convolutional neural networks to extract features and employed a switch map to adaptively fuse the predicted saliency maps.
		Chen~\cite{Chen2019Three} proposed a three-stream attention-aware multi-modal fusion network to improve the performance of saliency detection.
		Zhang et al.~\cite{Zhang2020UC} proposed a probabilistic RGB-D saliency detection model, which learns from the labeled data via conditional variational autoencoders.
		However, these methods usually employ simple concatenation or addition operations to aggregate RGB and depth features, which leads to unsatisfactory performance.
		In addition, the useless information are propagated, which degrades the saliency detection accuracy.
		
		To resolve these issues, we propose a novel fusion module to integrate cross-modal features.
		The proposed module utilizes weight coefficients learnt from lower level to enhance the details of RGB features at current level, which generates the complement RGB information to improve the model performance.
		The learned coefficients are then assigned to the RGB, complementary RGB and depth feature branches, which fuses the features adaptively to self-correction and yields improved saliency maps.
		Moreover, our module can improve the quality of salient maps and suppress the interferences in the complex or low-contrast scenes.
		
		\begin{figure*}[tb]
			\begin{center}
				\includegraphics[width=17.5cm]{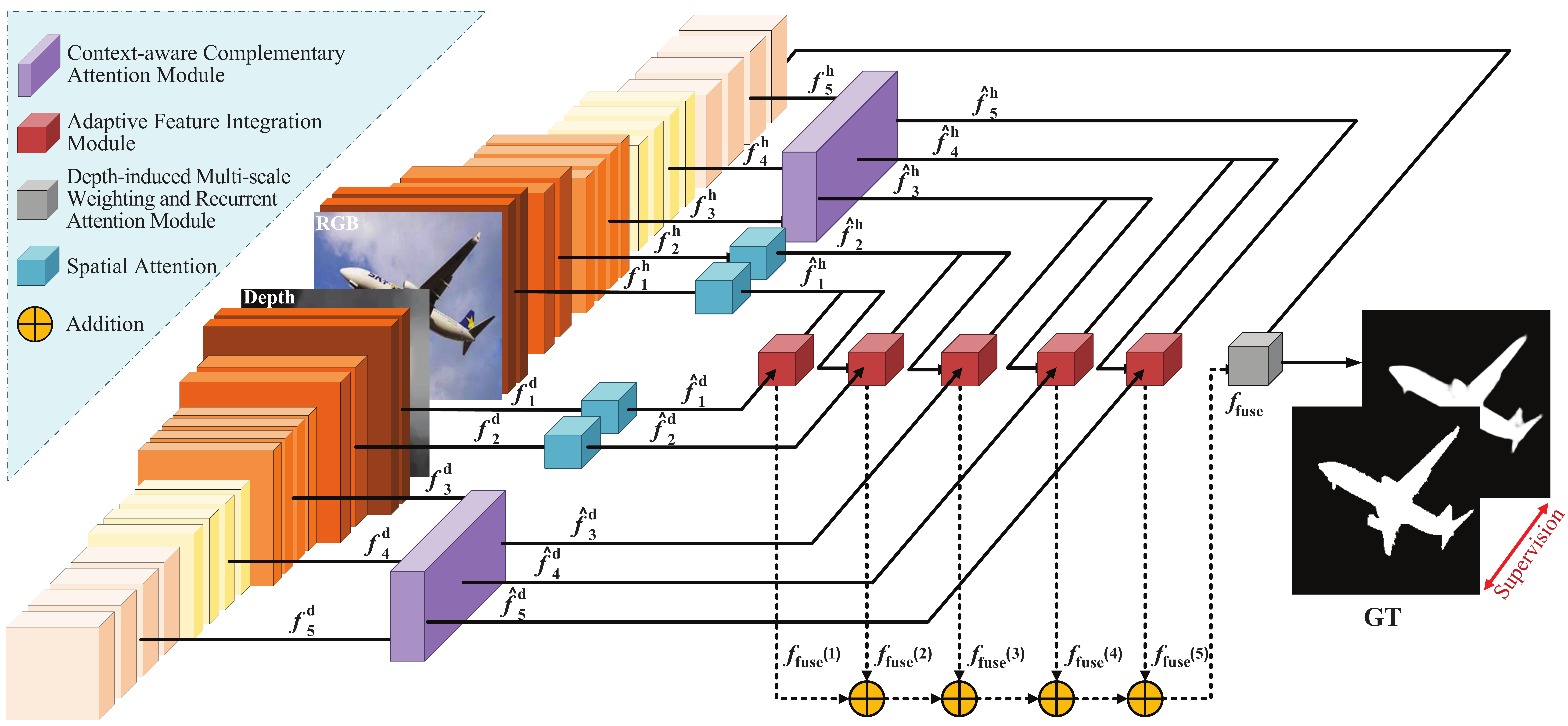}
			\end{center}
			\vspace{-5pt}
			\caption{An overview of our network. We propose a Complementary Attention and Adaptive Integration Network (CAAI-Net) with two modules, i.e., the context-aware complementary attention (CCA) module and adaptive feature integration (AFI) module.}
			\label{fig:all}
		\end{figure*}
		
		\subsection{Global and Local Context Mechanism}
		A number of studies have demonstrated that global and local information plays an important role in the effective salient object detection.
		Wang et al.~\cite{DBLP:conf/cvpr/WangZWL0RB18} proposed a global recurrent localization network, which exploits the weighted contextual information to improve accuracy of saliency detection.
		Liu et al.~\cite{Yihang2019Saliency} exploited the fusion of global and local information under multi-level cellular automata to detect saliency, and the global saliency map is obtained using the CNN-based encoder-decoder model.
		Ge et al.~\cite{Ge2019Saliency} obtained local information through superpixel segmentation, saliency estimation, and multi-scale linear combination. The resulting local information is fused with the CNN-based global information.
		Fu et al.~\cite{Fu2020JLDCF,fu2020siamese} proposed a joint learning and densely cooperative fusion architecture to acquire robust salient features.
		Chen et al.~\cite{DBLP:journals/corr/abs-2003-00651} proposed a global context-aware aggregation network, where a global module is designed to generate the global context information.
		The resulting context information is fused across different levels to compensate the missing information and to mitigate the dilution effect in high-level features.
		In this paper, local context features are acquired by a feature interaction component in the CCA module and then fed into a complementary attention component with the guidance from global context information to learn more meaningful features.

		\subsection{Attention Mechanism}
		The attention mechanism stems from the fact that human vision assigns more attention to the region of interests and suppresses the useless background information.
		Recently, it has been widely applied in various computer vision tasks~\cite{DBLP:journals/tip/WangS18,DBLP:conf/cvpr/WangSZSZHL19}.
		Li et al.~\cite{DBLP:conf/ijcai/LiCBH20} exploited the asymmetric co-attention to adaptively focus important information from different blocks at the interweaved nodes and to improve the discriminative ability of networks.
		Fu et al.~\cite{DBLP:conf/cvpr/FuLT0BFL19} proposed a dual attention network including position attention and channel attention module to capture long-range contextual information and to fuse local features with global features.
		Zhang et al.~\cite{Zhang2020Bilateral} introduced a bilateral attention module to capture more useful foreground and background cues and to optimize the uncertain details between foreground and background regions.
		Zhang et al.~\cite{DBLP:journals/corr/abs-2004-08955} presented a split-attention block to enhance the performance of learned features and to apply across vision tasks.
		Noori et al.~\cite{DBLP:journals/eaai/NooriMMBH20} adopted a multi-scale attention guided module and an attention-based multi-level integrator module to obtain more discriminative feature maps and assign different weights to multi-level feature maps.
		In our work, we suppress useless features and improve accuracy of salient object detection by our CCA module, which is based on the spatial attention and channel attention.
		
		\section{Method}\label{sec:method}
		In this section, we provide detail descriptions for the proposed RGB-D saliency detection model in terms of the overall network architecture and two major components, including CCA and AFI modules.
		Our network exploits the relationships between global and local features, high-level and low-level features, as well as different modality features.
		In addition, the features are fused effectively according to their respective characteristics.
		
		\subsection{Overall Architecture}
		Inspired by DMRANet~\cite{Yongri2020Depth}, the proposed network, CAAI-Net, considers both the global and local context information.
		Fig.~\ref{fig:all} shows an overview of CAAI-Net, which is based on a two-stream structure for RGB images and depth maps.
		As can be observed, CAAI-Net employs similar network branches to process the depth and RGB inputs.
		Low-level features have rich details, but the messy background information tends to affect the detection of salient objects.
		In contrast, high-level features have rich semantic information, which is useful for locating the salient objects, but the details are usually missing in the high-level features~\cite{Zhao2019Pyramid}.
		According to these characteristics, we divide the five convolutional blocks of VGG-19~\cite{Simonyan2014Very} into two parts, of which the first two convolution layers ($Conv1\_2$, $Conv2\_2$) are regarded as low-level features and the rest ($Conv3\_4$, $Conv4\_4$, $Conv5\_4$) are the high-level features.
		The high-level features are fed to our CCA module, which consists of three components (i.e., feature interaction component, complementary attention component, and global-context component), to obtain abundant context information and focus more on the regions with salient objects.
		The feature interaction component is proposed to extract sufficient features by fusing dense interweaved local context information.
		The output of feature interaction component is then fed into complementary attention component for extracting more meaningful features with the guidance of global context information.
		For the low-level features, we employ spatial attention components to refine them before the feature fusion. The underlying motivation lies in two folds. First, the attention mechanism has been demonstrated to be effective in improving the feature representation for capturing informative features, which is able to improve the performance effectively \cite{DBLP:journals/tip/WangS18,DBLP:conf/cvpr/WangSZSZHL19}.
		Second, as demonstrated by visualizing the features maps of CNNs \cite{DBLP:journals/corr/abs-2004-15004,DBLP:conf/eccv/ZeilerF14}, the low-level features contain abundant structural details (e.g., edges), indicating rich spatial information.
		Therefore, spatial attention components are employed to select effective features from the low-level features.
		We then utilize the AFI module to fuse the extracted RGB and depth features at all levels in an adaptive manner.
		Finally, the fused features at different levels are added together and then fed into the depth-induced multi-scale weighting and recurrent attention module~\cite{Yongri2020Depth} for predicting the saliency map.
		
		\begin{figure*}[t]
			\begin{center}
				\includegraphics[width=17.2cm]{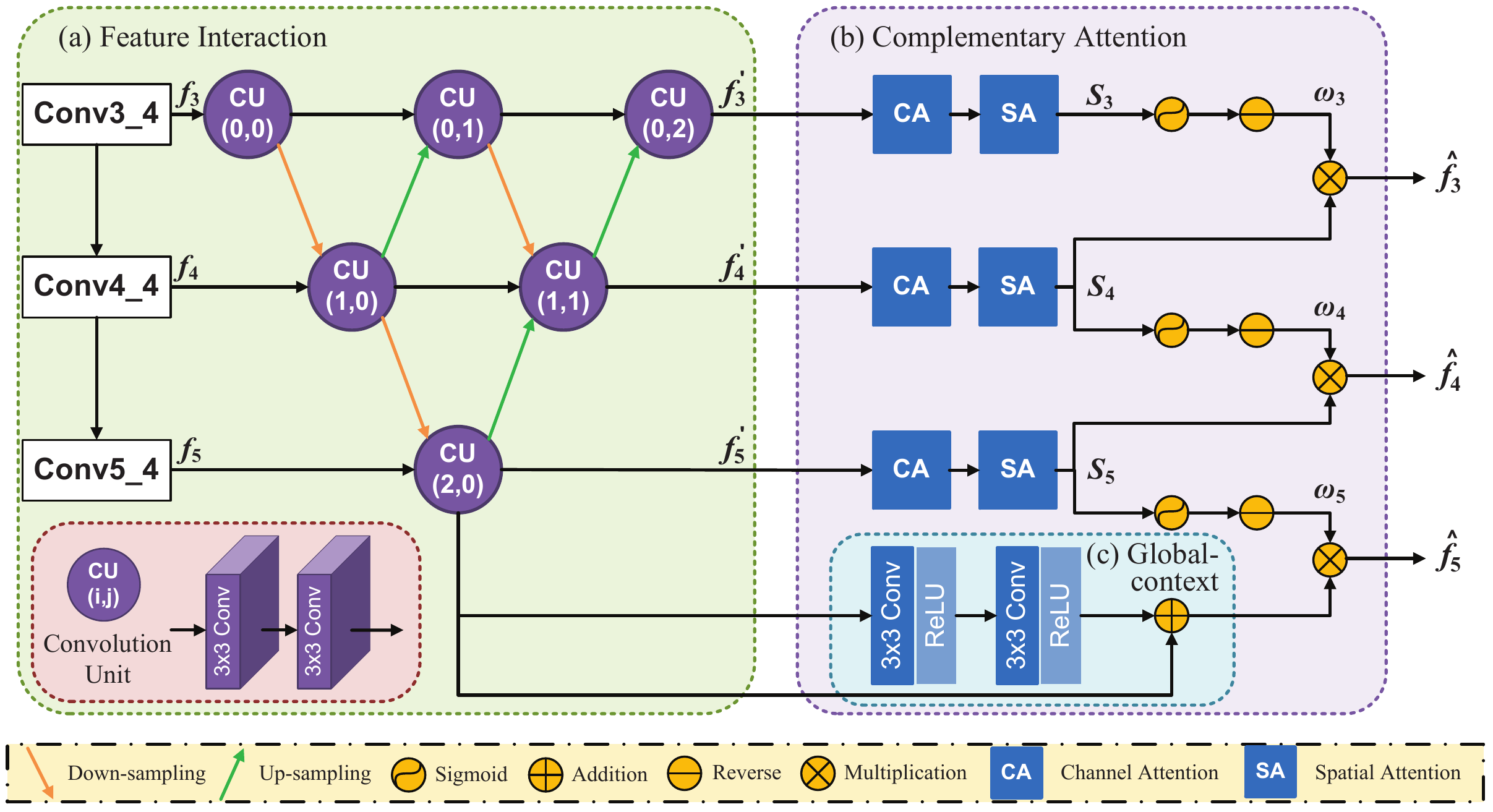}
			\end{center}
			\vspace{-5pt}
			\caption{Illustration of the Context-aware Complementary Attention (CCA) module. The CCA module consists of a feature interaction component, a complementary attention component, and a global-context component.}
			\label{fig:short}
		\end{figure*}
		\subsection{Context-aware Complementary Attention Module}
		An overview of our CCA module is shown in Fig.~\ref{fig:short}. We will then detail in its three major components as follows.
		
		\subsubsection{Feature Interaction Component}
		Extracting the local context information plays an important role in the task of RGB-D saliency detection.
		Previous works adopt various methods to obtain the local context information for capturing the informative features related to saliency detection.
		Liu et al.~\cite{Liu2018A} proposed a deep spatial contextual long-term recurrent convolutional network to boost the saliency detection performance by incorporating both global and local context information.
		Liu et al.~\cite{Yihang2019Saliency} employed a locality-constrained linear coding model to generate local saliency map by minimizing its reconstruction errors.
		Liu et al.~\cite{Nian2020PiCANet} proposed a pixel-wise contextual attention network to selectively focus on useful local-context information at each pixel, which can strengthen the performance of RGB-D saliency detection.
		
		A number of works have shown that combining the features of adjacent layers can more effectively supplement mutual features.
		Therefore, we design the feature interaction component for high-level features to capture the local context information across levels (see Fig.~\ref{fig:short} (a)).
		To suppress complex background information, we adopt the reticular pyramid to fuse multi-scale information, which yields the enhanced features $f_{i}^{'}$ with $i=3,4,5$. Note that we omit the superscripts, $\text{h}$ and $\text{d}$, for clarity.
		Mathematically, we define the feature interaction component as
			\begin{align}
			\label{eq:21}
			f_{(0,0)}&=CU_{(0,0)}(f_{3}),\\
			f_{(1,0)}&=CU_{(1,0)}(f_{4}+DS(f_{(0,0)})),\\
			f_{(0,1)}&=CU_{(0,1)}(f_{(0,0)}+US(f_{(1,0)})),\\
			f_{(2,0)}&=CU_{(2,0)}(f_{5}+DS(f_{(1,0)})),\\
			\label{eq:51}
			f_{(1,1)}&=CU_{(1,1)}(f_{(1,0)}+US(f_{(2,0)})+DS(f_{(0,1)})),\\
			f_{(0,2)}&=CU_{(0,2)}(US(f_{(1,1)})+f_{(0,1)}).
			\end{align}
		Taking Eq.~\eqref{eq:51} as an example, $f_{(1,1)}$ denotes the output of convolution unit $CU_{(1,1)}(\cdot)$. $US(\cdot)$ is the up-sampling operation via bilinear interpolation, and $DS(\cdot)$ is the down-sampling operation. $f_{i}$ with $i=3,4,5$ denotes the input of the $i$th layer.
		We then have the outputs of feature interaction component as 
		$f_{3}^{'}=f_{(0,2)}$, $f_{4}^{'}=f_{(1,1)}$, and $f_{5}^{'}=f_{(2,0)}$.
		Furthermore, the CPM can be extended to more layers, and the principle is similar to the three-layer pyramid structure in this paper.
		
		\subsubsection{Complementary Attention Component}
\begin{figure}[t]
			\begin{center}
				\includegraphics[scale=0.58]{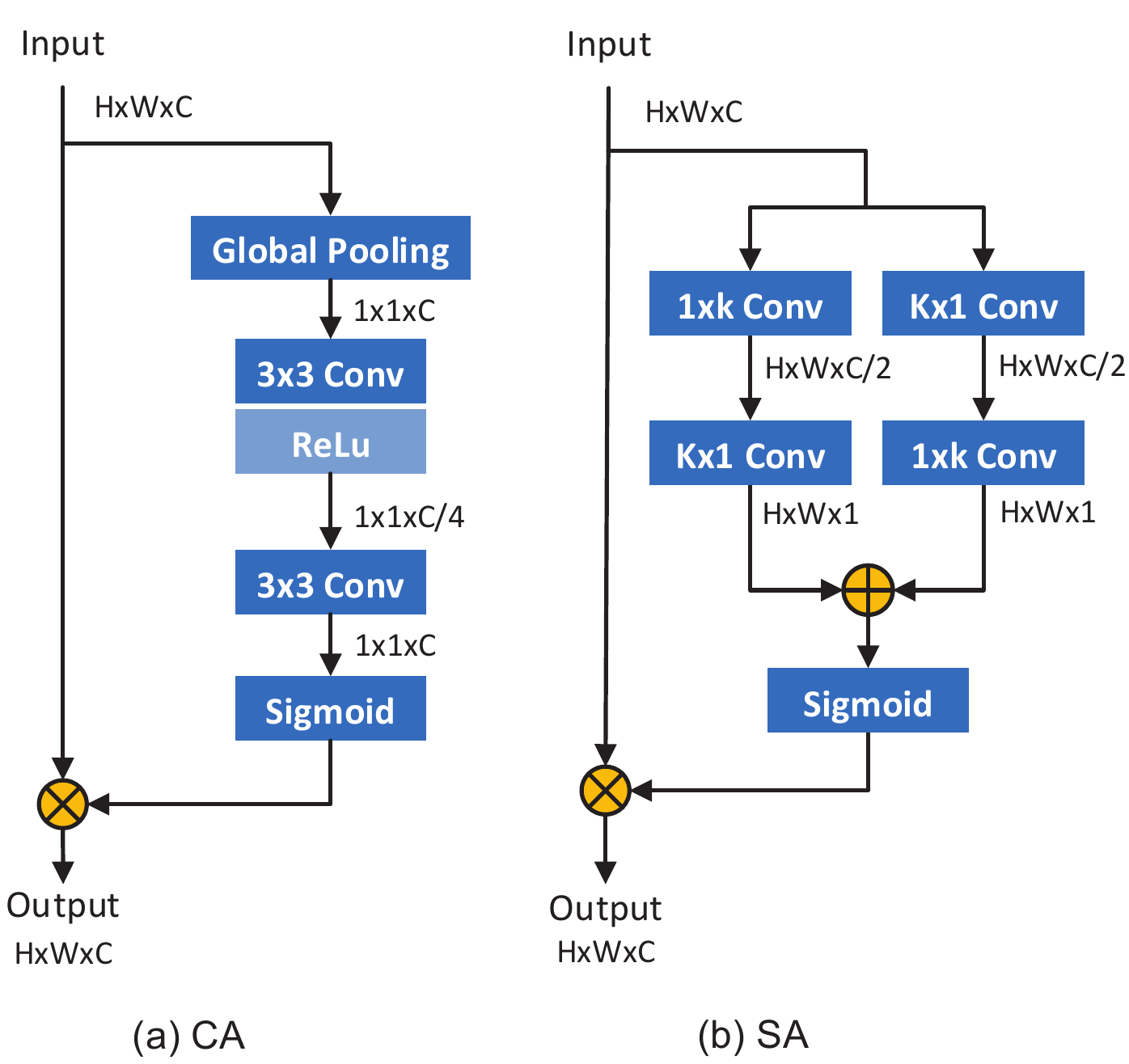}
			\end{center}
			\caption{Illustration of the channel attention (CA) and spatial attention (SA) components~\cite{Zhao2019Pyramid}.}
			\label{fig:CA}
		\end{figure}

		As shown in Fig.~\ref{fig:short} (b), in order to further reduce the background redundant information and locate interested regions, the outputs $f_{i}^{'}$ from the feature interaction component are fed into channel attention (see Fig.~\ref{fig:CA} (a)) and spatial attention (see Fig.~\ref{fig:CA} (b)) components~\cite{Zhao2019Pyramid}.
		Specifically, the features obtained from dual attention mechanism are first divided into two parts, one is the original output $S_{i}$, the other is a normalized and reversed one $\omega_{i}$, which is regarded as the weight factor learnt from supplementary attention for exploiting the interactive features between the adjacent levels.
		$\omega_{i}$ is then multiplied with the output $S_{i+1}$ of the next level to enhance the features and to supplement the details. Note that $k$ in the SA (see Fig.~\ref{fig:CA} (b)) component is taken as 5 to obtain the required size of output features.
		The first two outputs, $\hat{f}_{i}$ with $i=3,4$, of the CCA module are defined as
			\begin{equation}
			\hat{f}_{i}=\circleddash(\delta(S_{i}))\odot S_{i+1},
			\end{equation}
		where $\delta(\cdot)$ represents a Sigmoid activation function, $\odot$ denotes the Hadamard product, and $\circleddash(\cdot)$ represents a reverse operation \cite{chen2018reverse,fan2020inf}, which subtracts the input from a matrix of all ones.
		
		\subsubsection{Global-context Component}
		For the fifth-level features, global context information (see Fig.~\ref{fig:short} (c)) is introduced as the supplementary information to combine with the attention module, which is able to correct the location and enrich the features of salient objects.
		Simply adding the global with local features is not an effective solution, therefore we adopt the residual component as a rough locator to generate the global context information, i.e.,
			\begin{equation}
			\hat{f}_{5}=\omega_{5} \odot (f_{5}^{'}+\epsilon(Conv_{3\times3}(\epsilon(Conv_{3\times3}(f_{5}^{'})))))
			\end{equation}
		where $Conv_{3\times3}(\cdot)$ denotes a 3$\times$3 convolutional layer and $\epsilon(\cdot)$ denotes a ReLU activation function.
		\subsection{Adaptive Feature Integration Module}
		Although RGB and depth are complementary and depth can provide unique semantic information, the feature in the depth is not abundant in terms of structural details.
		If the depth information is treated equally with RGB, it may result in the degradation of model performance.
		Therefore, we develop AFI module, an effective fusion module, which is able to sufficiently integrate the features of the cross-modal to adaptively correct the impact of the depth features which have low-quality but abundant spatial information.
		
\begin{figure}[t]
			\begin{center}
				\includegraphics[scale=0.59]{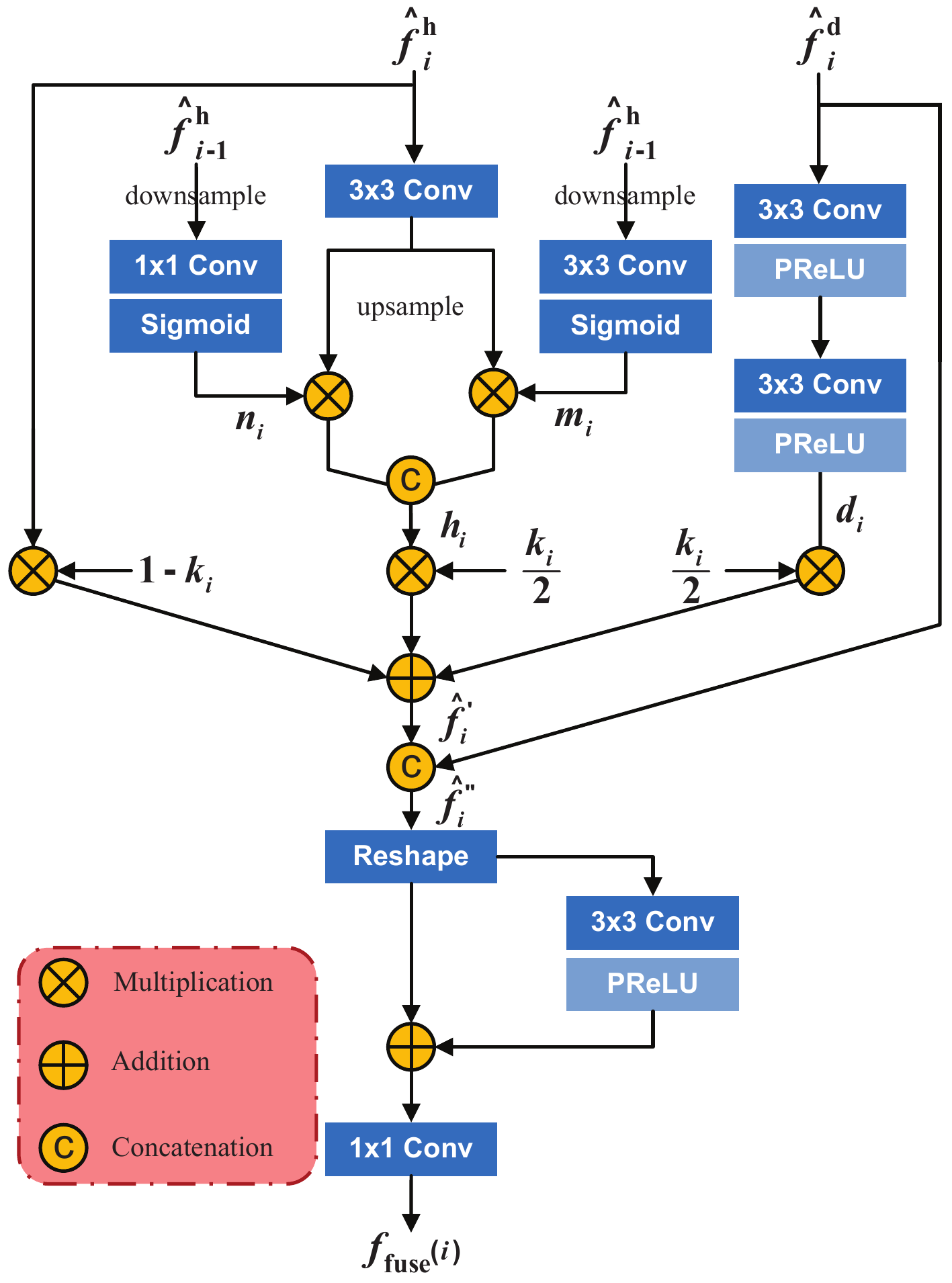}
			\end{center}
			\caption{Illustration of the adaptive feature integration (AFI) module.}
			\label{fig:FA}
		\end{figure}

		As illustrated in Fig.~\ref{fig:FA}, the inputs $\hat{f}_{i}^{\text{h}}$ and $\hat{f}_{i}^{\text{d}}$ with $i=1,2,3,4,5$ represent RGB and depth features at each layer, respectively.
		First, the RGB features of the lower layer are fed into 1$\times$1 convolution layer after down-sampling, so that the resulting features have the same number of channels as the higher-level features.
		Then, the correction factor $n_{i}$ is obtained using a Sigmoid layer.
		Moreover, taking different receptive fields into consideration, we apply a 3$\times$3 convolution layer to learn a balanced correction factor $m_{i}$.
		Further, these two symmetric weights are multiplied separately by the feature that is input into the 3$\times$3 convolution layer after up-sampling.
		They are then concatenated for the new features.
		In addition, to learn the depth feature $d_{i}$, $\hat{f}_{i}^{\text{d}}$ is fed into two units, each of which includes a convolutional layer followed by PReLU activation function. The depth map usually suffers from low-quality and noise issues, therefore treating depth and RGB features equally in the fusion leads to unsatisfactory results. To resolve this issue, we add the modified RGB features $h_{i}$, the depth features $d_{i}$, and the original RGB features $\hat{f}_{i}^{\text{h}}$ proportionally with a learned coefficient $k_{i}$, which is obtained using the RGB feature $\hat{f}_{i}^{\text{h}}$ and a pooling layer that reduces the feature dimension. We utilize the RGB information to guide the complementary and depth information so that the fused features provide a good representation of multi-modal features.
		Finally, the output is concatenated with the depth features $\hat{f}_{i}^{\text{d}}$. Mathematically, the above procedure is defined as
			\begin{align}
			n_{i}&=\delta(Conv_{1\times1}(DS(\hat{f}_{i-1}^{\text{h}}))),\\
			m_{i}&=\delta(Conv_{3\times3}(DS(\hat{f}_{i-1}^{\text{h}}))),\\
			h_{i}&=Cat(n_{i} \odot US(Conv_{3\times3}(\hat{f}_{i}^{\text{h}})),\nonumber\\ &~~~~~~~~~~~~~~~~~~~~~~~~~~~~m_{i} \odot US(Conv_{3\times3}(\hat{f}_{i}^{\text{h}}))),\\
			d_{i}&=\theta(Conv_{3\times3}(\theta(Conv_{3\times3}(\hat{f}_{i}^{\text{d}})))),\\
			\hat{f}_{i}^{'}&=(1-k_{i})\hat{f}_{i}^{\text{h}}+k_{i}(h_{i}+d_{i})/2,\\
			\hat{f}_{i}^{''}&=Cat(\hat{f}_{i}^{'},\hat{f}_{i}^{\text{d}}),
			\end{align}
		where $Conv_{1\times1}(\cdot)$ denotes a 1$\times$1 convolution layer. $Cat(\cdot)$ represents the concatenation operation, and $\theta(\cdot)$ denotes a PReLU activation function.
		
		\def \mycolor {green!75!black}
		\begin{table*}[]
			\centering
			\small
			\renewcommand{\arraystretch}{1.9}
			\caption{Quantitative results on six RGB-D benchmark datasets. Nine SOTA models are involved in the evaluation. The best three results are marked with \textcolor{red}{red}, \textcolor{\mycolor}{green} and \textcolor{blue}{blue} colors, respectively. Methods with/without ``$*$'' are trained with either the NJUD, NLPR, and DUT-RGBD training sets or the NJUD and NLPR training sets.
				``$\uparrow$'' indicates the higher the better, while ``$\downarrow$'' indicates the lower the better.}
			
			\setlength\tabcolsep{0.5pt}
			\resizebox{1\textwidth}{!}{
				\begin{tabular}{cc|cccc|cccc|cccc|cccc|cccc|cccc}
					\hline
					\rule{0pt}{20pt}\multirow{2}{*}{Methods}&\multirow{2}{*}{Years}&\multicolumn{4}{c|}{LFSD~\cite{Zhang2017Saliency}} &\multicolumn{4}{c|}{NJUD~\cite{Ran2015Depth}}&\multicolumn{4}{c|}{NLPR~\cite{Peng2014RGBD}}&\multicolumn{4}{c|}{STEREO~\cite{Niu2012Leveraging}}&\multicolumn{4}{c|}{RGBD135~\cite{DBLP:conf/icimcs/ChengFWXC14}}&\multicolumn{4}{c}{DUT-RGBD~\cite{Yongri2020Depth}} \\
					\cmidrule(r){3-6} \cmidrule(r){7-10} \cmidrule(r){11-14}\cmidrule(r){15-18} \cmidrule(r){19-22} \cmidrule(r){23-26}
					&&\rule{0pt}{10pt} $S_{\alpha}$ $\uparrow$       &   MAE $\downarrow$       &   maxE $\uparrow$    &    maxF$\uparrow$
					& $S_{\alpha}$ $\uparrow$       &   MAE $\downarrow$       &   maxE $\uparrow$   &    maxF $\uparrow$
					& $S_{\alpha}$ $\uparrow$       &   MAE $\downarrow$       &   maxE $\uparrow$   &    maxF $\uparrow$
					& $S_{\alpha}$ $\uparrow$       &   MAE $\downarrow$       &   maxE $\uparrow$   &    maxF $\uparrow$
					& $S_{\alpha}$ $\uparrow$        &   MAE  $\downarrow$     &   maxE $\uparrow$  &    maxF $\uparrow$
					& $S_{\alpha}$ $\uparrow$        &   MAE  $\downarrow$     &   maxE $\uparrow$  &    maxF $\uparrow$   \\
					\midrule	
					\multicolumn{1}{l}{MMCI~\cite{Chen2019Multi} }     &PR19      &0.787      &0.132  &0.839     &0.771        &0.859
					& 0.079    &0.915      &0.853  &0.856	&0.059	&0.913	&0.815
					&0.856       &0.080    &0.913    &0.843
					&0.848          &0.065    &0.928       &0.822       &0.791        &0.113
					&0.859			&0.767\\
					\multicolumn{1}{l}{TAN~\cite{Chen2019Three} }     &TIP19      & 0.801      & 0.111   &0.847   & 0.796       &0.878
					& 0.060     &0.925     & 0.874   &0.886	&0.041	&0.941	&0.863
					& 0.877       & 0.059    &0.927    &0.870
					& 0.858          & 0.046    &0.910       & 0.827       & 0.808        &0.093
					&0.861			&0.790\\
					\multicolumn{1}{l}{CPFP~\cite{zhao2019contrast}}      &CVPR19      & 0.828      & 0.088  &0.871    & 0.825    &0.878	&0.053	&0.923	&0.877
					&0.888	&0.036	&0.932	&0.868
					&0.879	&0.051	&0.925	&0.874
					& 0.874          & 0.037     &0.923      & 0.845       & 0.818        &0.076
					&0.859			&0.795\\
					
					\multicolumn{1}{l}{CFGA~\cite{Zhengyi2020A} }     &Neucom20  &0.802	&0.097	&0.858	&0.804	&0.885	&0.052	&0.925	&0.886	&\textcolor{\mycolor}{0.907}	&\textcolor{blue}{0.030}	&\textcolor{\mycolor}{0.948}	&\textcolor{\mycolor}{0.890}  &0.880	&0.050	&0.927	&0.879  &0.779	&0.061	&0.869	&0.709 &\textcolor{blue}{0.891}	&0.049	&0.923	&0.890\\
					
					\multicolumn{1}{l}{ASIF~\cite{Chongyi2020ASIF}}   &CVPR20      & 0.823      & 0.090   &0.860   & 0.824   &0.889      & \textcolor{\mycolor}{0.047}     &0.927     & 0.888          &\textcolor{blue}{0.906}	&\textcolor{blue}{0.030}	&0.944	&\textcolor{blue}{0.888} & 0.879       & 0.049   &0.927     &0.878
					& 0.764          & 0.076     &0.846      & 0.684       & 0.838        &0.073
					&0.876			&0.821\\
					
					\multicolumn{1}{l}{D3Net~\cite{fan2019rethinking} }     &TNNLS20     & 0.832      & 0.099  &0.864    & 0.819   &\textcolor{blue}{0.895}      & \textcolor{blue}{0.051}    &\textcolor{blue}{0.932}      & \textcolor{blue}{0.889}  &\textcolor{blue}{0.906}	&0.034	&\textcolor{blue}{0.946}	&0.886 & \textcolor{\mycolor}{0.901}       & 0.046    &\textcolor{\mycolor}{0.944}    &\textcolor{\mycolor}{0.898}
					& \textcolor{\mycolor}{0.906}          & 0.030     &\textcolor{blue}{0.939}      & \textcolor{blue}{0.882 }      & 0.814        &0.086 &0.857
					&0.786\\
					\hline
					\multicolumn{1}{l}{$^*$DMRA~\cite{Yongri2020Depth} }     &ICCV19      & 0.823      & \textcolor{blue}{0.087}  &\textcolor{blue}{0.886}   & \textcolor{blue}{0.841}    &0.880      & 0.053    &0.927      & \textcolor{blue}{0.889}   &0.890	&0.035	&0.940	&0.883  & 0.835       & 0.066   &0.911     &0.847
					& 0.878          & 0.035     &0.933      & 0.869       & 0.869        &0.057
					&0.927			&0.889\\
					
					\multicolumn{1}{l}{$^*$A2dele~\cite{piao2020A2dele}}      &CVPR20      & \textcolor{blue}{0.837}      & \textcolor{\mycolor}{0.074}  &0.880    & 0.836      &0.869      & \textcolor{blue}{0.051}  &0.916        & 0.873    &0.896	&\textcolor{\mycolor}{0.028}	&0.945	&0.880
					&0.885       & \textcolor{\mycolor}{0.043}    &0.935    &0.885
					&0.885          & \textcolor{blue}{0.028}     &0.923      & 0.867       &0.885       &\textcolor{blue}{0.042} &\textcolor{blue}{0.930}
					&\textcolor{blue}{0.892}\\
					
					\multicolumn{1}{l}{$^*$SSF~\cite{zhang2020Select}}      &CVPR20      & \textcolor{\mycolor}{0.859}      & \textcolor{red}{0.066}  &\textcolor{\mycolor}{0.900}    & \textcolor{\mycolor}{0.866}     &\textcolor{\mycolor}{0.899}      & \textcolor{red}{0.043}    &\textcolor{\mycolor}{0.935}      & \textcolor{\mycolor}{0.896}   &0.888	&0.035	&0.934	&0.864 & \textcolor{blue}{0.893}       & \textcolor{blue}{0.044}    &\textcolor{blue}{0.936}    &\textcolor{blue}{0.890}
					& \textcolor{blue}{0.905}          & \textcolor{red}{0.025}     &\textcolor{\mycolor}{0.941}      & \textcolor{\mycolor}{0.883}       & \textcolor{\mycolor}{0.915}        &\textcolor{red}{0.033} &\textcolor{\mycolor}{0.951}
					&\textcolor{\mycolor}{0.924}\\
					\hline
					\multicolumn{1}{l}{Ours}     &--      & \textcolor{red}{0.866}      & \textcolor{red}{0.066}   &\textcolor{red}{0.906}   & \textcolor{red}{0.867}        &\textcolor{red}{0.903}
					& \textcolor{red}{0.043}     &\textcolor{red}{0.940}     &\textcolor{red}{0.905}    &\textcolor{red}{0.912}	&\textcolor{red}{0.027}	&\textcolor{red}{0.949}	&\textcolor{red}{0.897}
					& \textcolor{red}{0.902}      & \textcolor{red}{0.041}   &\textcolor{red}{0.945}     &\textcolor{red}{0.902}
					& \textcolor{red}{0.909}          & \textcolor{\mycolor}{0.026}    &\textcolor{red}{0.946}       & \textcolor{red}{0.900}       & \textcolor{red}{0.916}      &\textcolor{\mycolor}{0.035}
					&\textcolor{red}{0.953}			&\textcolor{red}{0.927}\\	
					\hline		
			\end{tabular}}
			\label{tab:Compare}
		\end{table*}
		
		Furthermore, the output $\hat{f}_{i}^{''}$ is fed into the traditional residual unit to obtain the cross-modal fused feature $f_{\text{fuse}}(i)$ at each layer.
		Finally, the features at different layers are added to obtain the final features $f_{\text{fuse}}$, i.e.,
			\begin{equation}
			f_{\text{fuse}}=\sum_{i=1}^{5}f_{\text{fuse}}(i),
			\end{equation}
		
		Our AFI module allows RGB and depth information to be effectively fused according to their own characteristics in order to improve the saliency detection performance.
		
		\section{Experiments}\label{sec:experiments}
		In this section, we first introduce the implementation details, datasets, and evaluation metrics.
		We then present the experimental results to demonstrate the effectiveness of the proposed model by comparing with the SOTA models.
		Finally, we perform ablation analysis to investigate the proposed components.
		
		\subsection{Implementation Details}
		The proposed model is implemented using PyTorch, and the input images for training and testing are resized to 256$\times$256 before feeding into the network.
		The batch size is set to 2 and the training is optimized by mini-batch stochastic gradient descent.
		Other parameter settings are as follows: Learning rate is set to 1e-10, the momentum is set to 0.99, and the weight decay is set to 0.0005.
		Our model takes 61 epochs to complete the training.
		
		\subsection{Datasets}
		We evaluate the proposed method on six public RGB-D saliency detection benchmark datasets, which are detailed as follows:
		LFSD~\cite{Zhang2017Saliency} includes 100 RGB-D images and the depth maps are collected by Lytro camera.
		NJUD~\cite{Ran2015Depth} is composed of 1985 RGB images and corresponding depth images estimated from the stereo images with various objects and complex scenes.
		NLPR~\cite{Peng2014RGBD} consists of 1000 RGB images and corresponding depth images captured by Kinect.
		STEREO~\cite{Niu2012Leveraging} contains 797 stereoscopic images captured from the Internet.
		RGBD135~\cite{DBLP:conf/icimcs/ChengFWXC14} contains 135 RGB-D images captured by Kinect.
		DUT-RGBD~\cite{Yongri2020Depth} consists of 1200 paired images containing more complex real scenarios by Lytro camera.
		
		\subsection{Evaluation Metrics}
		Four evaluation metrics widely used in the field of RGB-D saliency detection are adopted in our experiments.
		These metrics include Structure Measure (S-Measure)~\cite{FanStructMeasureICCV17}, Mean Absolute Error (MAE)~\cite{Fitzgibbon2012}, E-measure~\cite{Fan2018Enhanced} and F-Measure~\cite{arbelaez2011contour}, each of which is detailed as follows.

		1) Structure Measure ($S_{\alpha}$)~\cite{FanStructMeasureICCV17}: This is a evaluation metric to measure the structural similarity between the predicted saliency map and the ground-truth map.
		According to~\cite{FanStructMeasureICCV17}, $S_{\alpha}$ is defined as
		\vspace{-4pt}
		\begin{equation}
		S_{\alpha}= \left ( 1-\alpha  \right )S_{o}+\alpha S_{r},
		\end{equation}
		where $S_{o}$ denotes the object-aware structural similarity and $S_{r}$ denotes the region-aware structural similarity.
		Following~\cite{FanStructMeasureICCV17}, we set $\alpha=0.5$.
		Note that the higher the S-measure score, the better the model performs.
		
		2) Mean absolute error ($MAE$)~\cite{Fitzgibbon2012}: This is a metric to directly calculate the average absolute error between the predict saliency map and the ground-truth.
		$MAE$ is defined as
		\begin{equation}
		MAE=\frac{1}{H\times W}\sum_{x=1}^{W}\sum_{y=1}^{H}\left | S\left ( x,y \right )-G\left ( x,y \right ) \right |
		\end{equation}
		where $H$ and $W$ denotes the height and width of the saliency map, respectively.
		$S$ represents the predicted saliency map, and $G$ denotes the corresponding ground truth. $x$ and $y$ denote the coordinate of each pixel.
		Note that the lower the $MAE$, the better the model performance.
		
		3) F-measure ($F_{\beta}$)~\cite{arbelaez2011contour}: This metric represents the weighted harmonic mean of recall and precision under a non-negative weights $\beta$.
		In the experiments, we use the maximum F-Measure ($maxF$) to evaluate the model performance.
		Mathematically, $F_{\beta}$ is defined as
		\begin{equation}
		F_{\beta }=\frac{\left ( \beta ^{2}+1 \right )Precision\times Recall}{\beta ^{2}Precision+Recall}
		\end{equation}
		Following~\cite{Zhang2017Amulet}, we set $\beta^{2}=0.3$.
		Note that the higher the F-measure score, the better the model performs.
		
		4) E-measure~\cite{Fan2018Enhanced}: E-measure is a perceptual-inspired metric and is defined as
		\begin{equation}
		E=\frac{1}{W\times H}\sum_{x=1}^{W}\sum_{y=1}^{H}\phi _{FM}\left ( x,y \right )
		\end{equation}
		where $\phi_{FM}$ is an enhanced alignment matrix~\cite{Fan2018Enhanced}.
		We adopt maximum E-Measure ($maxE$) to assess the model performance.
		Note that the higher the E-measure score, the better the model performs.
		\begin{figure*}[t]
			\centering
			\includegraphics[scale=0.63]{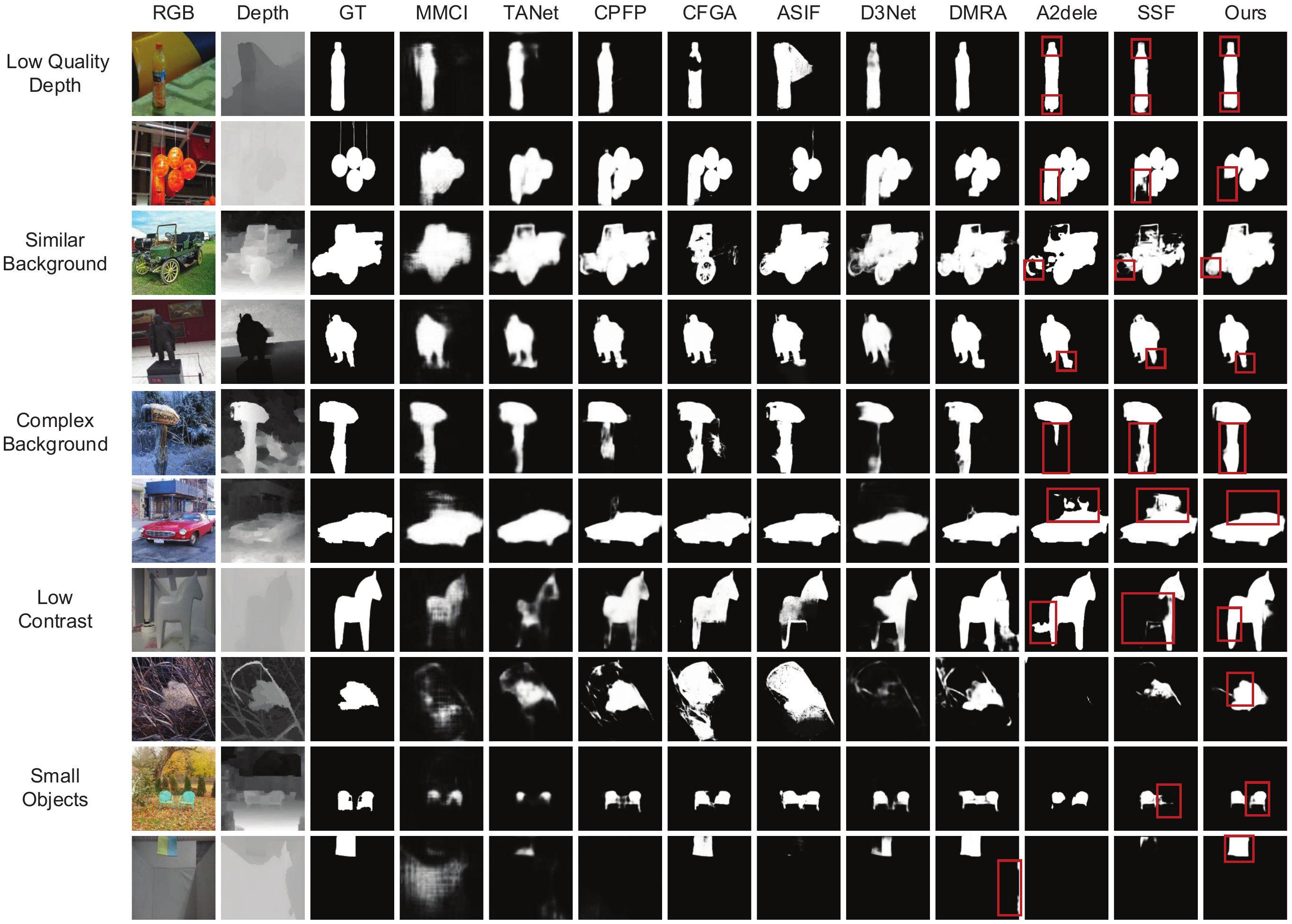}
			\caption{Qualitative visual comparison of the proposed model and the state-of-the-art models. Our model yields results that are closer to the ground truth maps than other models.}
			\label{fig:re}
		\end{figure*}
		
		\subsection{Comparison with State-of-the-arts}
		We perform extensive experiments to compare our CAAI-Net with nine state-of-the-art RGB-D saliency detection models, including DMRA~\cite{Yongri2020Depth}, CPFP~\cite{zhao2019contrast}, MMCI~\cite{Chen2019Multi}, TAN~\cite{Chen2019Three}, CFGA~\cite{Zhengyi2020A}, A2dele~\cite{piao2020A2dele}, SSF~\cite{zhang2020Select}, ASIF-Net~\cite{Chongyi2020ASIF} and D3Net~\cite{fan2019rethinking}. For fair comparison, we adopt the results provided by the authors directly or generate the results using the open source codes with default parameters. 
		In addition, for models without the source code publicly available, we adopt the corresponding published results.
		Our model is trained using the same training set with~\cite{Yongri2020Depth,piao2020A2dele,zhang2020Select}, which contains 800 samples from the DUT-RGBD, 1485 samples from NJUD and 700 samples from NLPR datasets.
		The remaining images in these datasets and other three datasets are used for testing.
		
		\textbf{Quantitative evaluation.}
		The results, shown in Table~\ref{tab:Compare}, indicate that CAAI-Net achieves promising performance on all six datasets and outperforms the SOTA models.
		Specifically, CAAI-Net sets new SOTA in terms of $S_{\alpha}$, $maxF$ and $maxE$ on all datasets.
		In addition, it provides the best $MAE$ results on four benchmark datasets and the second best $MAE$ results on RGB135 and DUT-RGBD. On the NLPR dataset, our model outperforms the second best with 3.8$\%$ improvement on $maxF$. 
		It is worth noting that CAAI-Net outperforms SOTA models on the DUT-RGBD and STEREO, which are challenging datasets that are with complex background information.
		All the quantitative results demonstrate that CAAI-Net is capable of improving the performance effectively.
		
		\begin{table}[t!]
			\centering
			\renewcommand\arraystretch{1.5}
			\caption{Ablation study of the proposed model.
				The best results are \textbf{bold}. ``$\uparrow$'' indicates the higher the better, while ``$\downarrow$'' indicates the lower the better.
				``B'' denotes the backbone module.
				``B+CCA'' denotes the model with backbone and CCA module. ``B+CCA+AFI'' represents the model with backbone, CCA and AFI module.}
			\setlength{\tabcolsep}{0.2mm}{
				\resizebox{0.48\textwidth}{!}{
					\begin{tabular}{c|cccc|cccc|cccc}
						\hline
						\rule{0pt}{17pt}\multirow{2}{*}{Methods}  & \multicolumn{4}{c|}{NJUD} &\multicolumn{4}{c|}{STEREO}&\multicolumn{4}{c}{DUT-RGBD}\\  \cmidrule(r){2-5} \cmidrule(r){6-9} \cmidrule(r){10-13}
						& $S_{\alpha}$ $\uparrow$       &   MAE $\downarrow$  &  maxE$\uparrow$  &  maxF$\uparrow$
						&$S_{\alpha}$ $\uparrow$       &   MAE $\downarrow$  &  maxE$\uparrow$  &  maxF$\uparrow$
						& $S_{\alpha}$ $\uparrow$       &   MAE $\downarrow$  &  maxE$\uparrow$  &  maxF$\uparrow$\\
						\hline
						\multicolumn{1}{l|}{B} &0.88  &0.053 &0.927 &0.889  &0.835  &0.066 &0.911 &0.847  &0.869  &0.057 &0.927 &0.889\\
						\multicolumn{1}{l|}{B+CCA} &0.900  &0.044  &0.937 &0.896  &0.898  &0.044 &0.943 &0.897  &0.912  &0.036 &\textbf{0.953} &0.921\\
						\multicolumn{1}{l|}{B+CCA+AFI}  &\textbf{0.903}  &\textbf{0.043}   &\textbf{0.94} &\textbf{0.905}  &\textbf{0.902}  &\textbf{0.041} &\textbf{0.945} &\textbf{0.902}  &\textbf{0.916}  &\textbf{0.035}  &\textbf{0.953}
						&\textbf{0.927}\\
						\hline	
						
			\end{tabular}}}
			\label{tab:Ablation}
		\end{table}	
		
		\textbf{Qualitative evaluation.}
		We further show the visual comparison of predicted saliency maps in Fig.~\ref{fig:re}.
		As can be observed, CAAI-Net yields saliency maps that are close to the ground truth. In contrast, the competing methods provide unsatisfactory results that poses significant differences with the ground truth.
		In particular, for the challenging cases, such as low-quality depth, background interference, low contrast, and small objects, CAAI-Net consistently provides promising results and outperforms the competing methods significantly.
		Specifically, the first two rows of Fig.~\ref{fig:re} shows the results for the case of low-quality depth.
		Although challenging, CAAI-Net overcomes the low-quality issue and accurately detects the salient objects, especially for the regions marked by red rectangles.
		Besides, the object and the background have similar colors in the next two rows.
		The next two rows show the case of similar background where the salient object shares similar appearance with the background.
		Our model consistently provides the best performance in comparison with competing methods.
		The results, shown in the fifth and sixth rows, indicate that CAAI-Net consistently provides the best performance in the presence of complex background problems.
		Finally, the last four rows show the resulting regarding low contrast and small objects.
		The effectiveness of our method is further confirmed by these two challenging cases.
		
		\begin{figure}[tb]
			\begin{center}
				\includegraphics[scale=0.52]{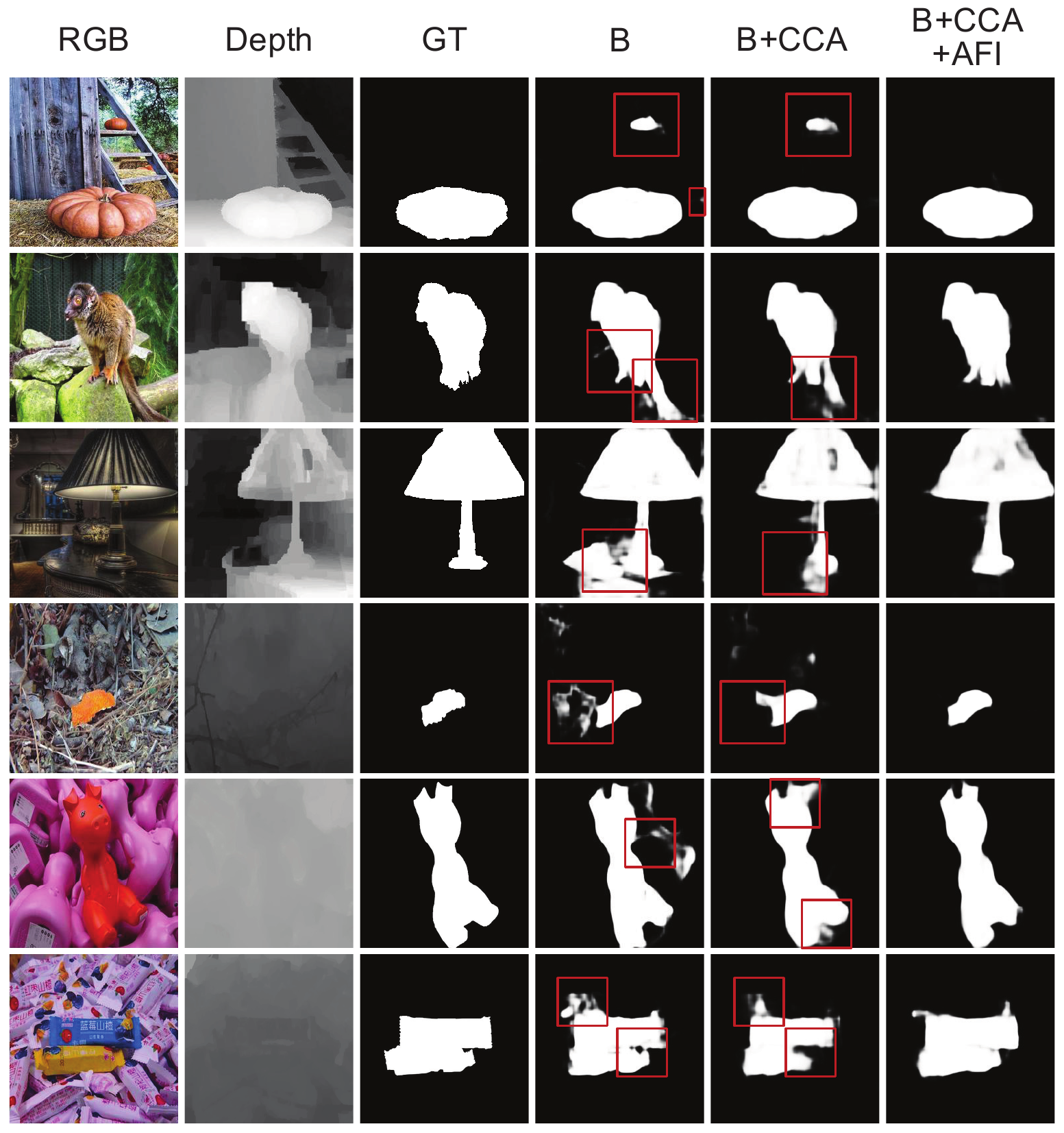}
			\end{center}
			\caption{Visual comparison of the ablated versions of our model. ``B'' denotes the backbone module. ``B+CCA'' denotes the model with backbone and CCA module. ``B+CCA+AFI'' means the model with backbone, CCA and AFI module.}
			\label{fig:ab}
		\end{figure}

		\subsection{Ablation Study}
		In this section, the ablation experiments on three testing datasets are performed to validate the effectiveness of the proposed CCA and AFI modules.
		
		\textbf{Effectiveness of CCA module.}
		The results, shown in Table~\ref{tab:Ablation}, indicate that the ablated version, B+CCA, outperforms the backbone network, B, in all datasets and evaluation metrics, demonstrating that the CCA module is an effective module to improve the performance.
		In particular, CCA module significantly reduces the MAE value, indicating that the predicted saliency maps are much closer to the ground truth.
		The advantage of CCA module can be attributed to its ability of locating the interested regions more accurately.
		In addition, the visual results, shown in Fig.~\ref{fig:ab}, provides the consistent conclusion, as in Table~\ref{tab:Ablation}. Our CCA module is an effective module for improving the accuracy of saliency detection.
		
		\begin{table}[t!]
			\centering
			\renewcommand\arraystretch{1.5}
			\caption{Ablation study of CCA module. ``B'' denotes the baseline module without three components, i.e., $g_i = f_i$ with $i=3,4,5$. ``B+(a)'' denotes the module with feature interaction component. ``B+(a)+(b)'' represents the module with feature interaction and complementary attention components. ``B+(a)+(b)+(c)'' denotes the module with feature interaction, complementary attention, and global-context components.}
			\setlength{\tabcolsep}{0.2mm}{
				\resizebox{0.48\textwidth}{!}{
					\begin{tabular}{c|cccc|cccc|cccc}
						\hline
						\rule{0pt}{17pt}\multirow{2}{*}{Methods}  & \multicolumn{4}{c|}{NJUD} &\multicolumn{4}{c|}{STEREO}&\multicolumn{4}{c}{DUT-RGBD}\\  \cmidrule(r){2-5} \cmidrule(r){6-9} \cmidrule(r){10-13}
						& $S_{\alpha}$ $\uparrow$       &   MAE $\downarrow$  &  maxE$\uparrow$  &  maxF$\uparrow$
						&$S_{\alpha}$ $\uparrow$       &   MAE $\downarrow$  &  maxE$\uparrow$  &  maxF$\uparrow$
						& $S_{\alpha}$ $\uparrow$       &   MAE $\downarrow$  &  maxE$\uparrow$  &  maxF$\uparrow$\\
						\hline
						\multicolumn{1}{l|}{B} &0.88  &0.053 &0.927 &0.889  &0.835  &0.066 &0.911 &0.847  &0.869  &0.057 &0.927 &0.889\\
						\multicolumn{1}{l|}{B+(a)} &0.898  &0.048 &0.935  &\textbf{0.897}  &0.895  &0.048 &0.939 &0.892  &0.904  &0.043 &0.946 &0.911\\						
						\multicolumn{1}{l|}{B+(a)+(b)}  &0.895  &0.048   &0.931 &0.895  &\textbf{0.898}  &0.045 &0.942 &\textbf{0.898}  &0.907  &0.041  &0.946
						&0.916\\
						\multicolumn{1}{l|}{B+(a)+(b)+(c)} &\textbf{0.900}  &\textbf{0.044}  &\textbf{0.937} &0.896  &\textbf{0.898}  &\textbf{0.044} &\textbf{0.943} &0.897  &\textbf{0.912}  &\textbf{0.036} &\textbf{0.953} &\textbf{0.921}\\
						
						\hline	
						
			\end{tabular}}}
			\label{tab:CCAabl}
		\end{table}	
		
		In addition, we further investigate the effectiveness of each component of CCA module by performing ablation studies. The results, shown in Table~\ref{tab:CCAabl}, indicate that ``B+(a)'' outperforms the baseline module ``B'' across different datasets, sufficiently demonstrating the effectiveness of our feature interaction component. The results, shown in the third row of Table~\ref{tab:CCAabl}, indicate that the complementary attention component effectively improves the performance on the complex scene (i.e., STEREO and DUT-RGBD). The complementary attention component enables the model to put more emphasis on informative features and suppressing background interferences.
		Finally, we show the results for the full version of CCA in the fourth row of Table~\ref{tab:CCAabl}. As can be observed, the global-context component improves the performance effectively, demonstrating its advantages.
		
		\textbf{Effectiveness of AFI module.}
		We then investigate the effectiveness of AFI module.
		The results, shown in Table~\ref{tab:Ablation}, indicate that the full version of our model with AFI module outperforms the ablated version, B+CCA, in terms of all evaluation metrics.
		This sufficiently demonstrates the effectiveness of AFI, which is capable of adaptively fusing the multi-modal features to capture the meaningful features for accurate saliency detection.
		In addition, the visual results, shown in Fig.~\ref{fig:ab}, confirm our observation in Table~\ref{tab:Ablation}, further demonstrating the effectiveness of AFI module sufficiently.
		As can be observed, the full version of our model yields saliency maps that are close to the ground truth. In contrast, B+CCA fails to provide satisfactory results, especially in the regions marked by rectangles.

		\subsection{Failure Cases}
\begin{figure}[t]
			\begin{center}
				\includegraphics[scale=0.65]{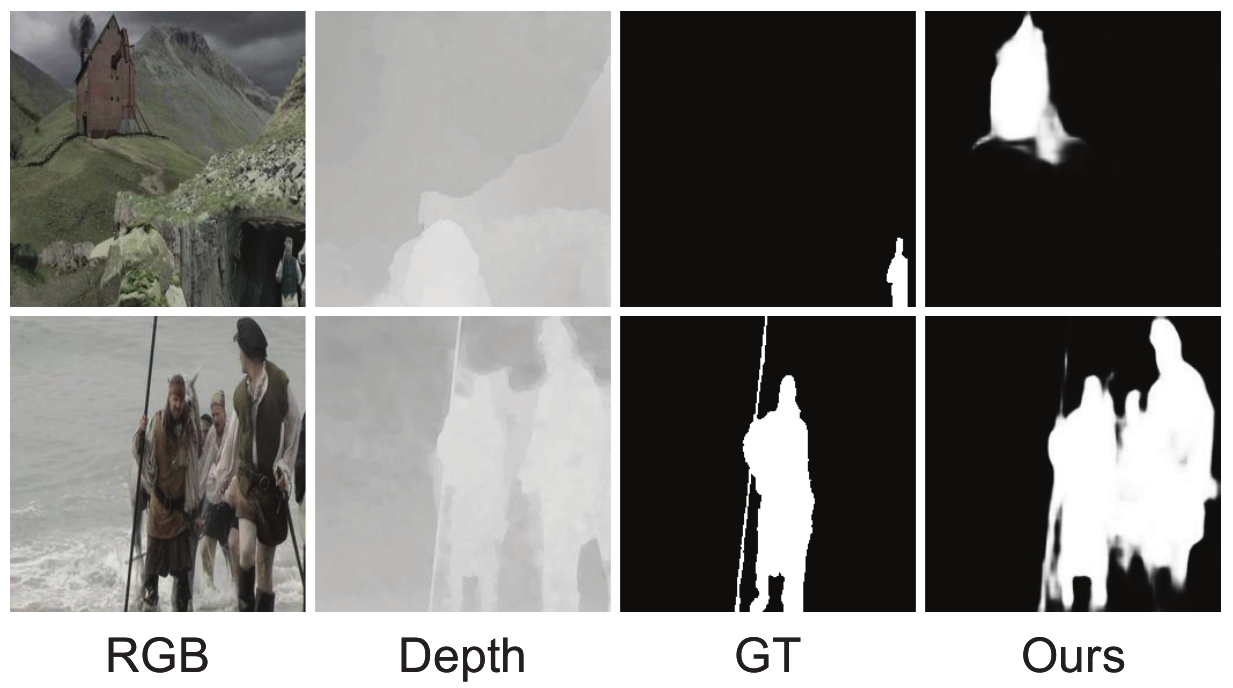}
			\end{center}
			\vspace{-12pt}
			\caption{Failure cases of CAAI-Net in extreme scenarios.}
			\label{fig:ff}
		\end{figure}

		Despite its various advantages, our model may yield mis-detections for some extreme scenarios. For instance, as shown in the top row of Fig.~\ref{fig:ff}, the object in image background is recognized as the salient one by mistake. In addition, as shown in the bottom row of Fig.~\ref{fig:ff}, the detection accuracy decreases when the background objects share similar appearances with the target salient object. {In the future, we will consider more comprehensive scenarios and explore more effective solutions to handle these challenging saliency detection tasks.}
		
		\section{Conclusion}\label{sec:conclusion}
		In this paper, we have proposed a novel RGB-D saliency detection network, CAAI-Net, which extracts and fuses the multi-modal features effectively for accurate saliency detection.
		Our CAAI-Net first utilizes the CCA module to extract informative features highly related to the saliency detection.
		The resulting features are then fed to our AFI module, which adaptively fuses the cross-modal features according to their contributions to the saliency detection.
		Extensive experiments on six widely-used benchmark datasets demonstrate that CAAI-Net is an effective RGB-D saliency detection model and outperforms cutting-edge models, both qualitatively and quantitatively.
		
		\section*{Declaration of Competing Interest}
		The authors declare that they have no known competing financial interests or personal relationships that could have appeared to
		influence the work reported in this paper.
		
		
		\bibliographystyle{elsarticle-num}
		
		\bibliography{cas-refs}

	\end{sloppypar}
\end{document}